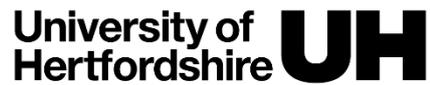

University of Hertfordshire
Hatfield, 2025

School of Physics, Engineering and Computer Science

MSc Artificial Intelligence and Robotics
7COM1039-0206-2024
Advanced Computer Science Master's Project
This report has been critically proofread and quality checked

*Evaluating Reinforcement Learning Algorithms for Navigation in Simulated Robotic Quadrupeds: A Comparative Study Inspired by Guide Dog Behaviour*

Student: Emma M. A. Harrison

ID Number: 23040079

Supervisor: Dr. Diego Resende Faria

Date: 27/02/2025

**Abstract**

Robots are found to be integrated in many industries, especially within the healthcare sector. However, many valuable applications, specifically for quadrupedal robots, are overlooked. This research explores the effectiveness of three different types of reinforcement learning algorithms in training a simulated quadruped robot for autonomous navigation and obstacle avoidance.

The goal is to develop a robotic guide dog simulation capable of following a path and avoiding obstacles. The long-term aim is to assess whether a real-world robotic quadruped could assist guide dogs and visually impaired individuals. The objective is also to expand the research horizons in medical 'pets', including, but not limited to, robotic guide dogs and alert dogs.

A comparative analysis of thirteen related research papers has been extensively studied, which helped define key evaluation criteria, including collision detection, path-finding algorithms, sensor usage, robot type, and simulation platforms. The research will focus on data collection from sensor inputs, collision frequency, episode/rewards, and learning rate over time to determine which algorithm performs best for robotic navigation in complex and dynamic environments.

The environments tested in this paper are custom-made to ensure fairness of all three algorithms being tested in a controlled environment. Data can be collected and compared equally across the controlled environments to evaluate the most successful algorithm honestly.

Results show that Proximal Policy Optimization (PPO) outperformed all other algorithms (Deep Q-Network and Q-learning), in all key areas recorded, as was expected. A result that significantly expresses this could be the average and median steps to the goal per episode for PPO, DQN, and Q.

By analysing these results, this study aims to contribute to the field of robotic navigation and medical robotics, providing insights into the feasibility of AI-driven quadruped mobility and its potential role in assistive robotics.

**Acknowledgements**

I wish to express my thanks and gratitude to my partner, Dylan, and good friend, Ross, who have kept me driven, provided support and encouragement, and have been the foundations of my academic success. I also want to thank my mother, who has given me constant love and moral support and continuously reminded me that I can achieve this.

I am also grateful to Dr. Diego Resende Faria, who has continued to steer me in the right direction and push my academia to the max. I am also grateful for his continued support in helping to get this paper published in a Q1 journal.

Finally, I express my thanks for God's continual blessings and the strength I have received to allow me to keep going.

**MSc Final Project Declaration**

This report is submitted in partial fulfillment of the requirements for the degree of:
Master of Science in Advanced Computer Science, Master's Project, [7COM1039-0206-2024] at the University of Hertfordshire (UH). I hereby declare that the work presented in this project and report is entirely my own, except where explicitly stated otherwise. All sources of information and ideas, whether quoted directly or paraphrased, have been properly referenced in accordance with academic standards. I understand that any failure to properly acknowledge the work of others could constitute plagiarism and may result in academic penalties. I did not use human participants in my MSc Project. I hereby give permission for the report to be made available on the university website, provided the source is acknowledged.

# Table of Contents

# 1. Introduction

## 1.1. Context and Motivation

As new innovative technologies advance in the robotics industry, it is vital to continue research and development to revolutionise the industry further. This paper addresses the efficacy of reinforcement learning algorithms for navigation in simulated robotic quadrupeds: A comparative study inspired by guide dog behaviour.

The purpose is to evaluate which algorithm is most effective for a quadruped when navigating simple and complex, dynamic environments carefully and safely. This would allow us to assess whether robot dogs could match the complexity of training required to that of a trained guide dog. This paper will aim to address which algorithm performs best regarding navigation efficiency in different environments. Do different sensor setups influence path planning performance? Also, how does the need for training reinforcement learning algorithms, such as Q-learning, compare to the immediate usability of deeper layered networks in DQN and the Actor-Critic Framework in PPO? PPO is expected to outperform other algorithms, especially in dynamic environments. It is also likely that with further time and research, quadrupedal robots will be able to perform equally as well as true guide dogs regarding navigation and obstacle avoidance. It is essential to understand and stay informed about the effects that robotics has on all industries, especially when working with robotics in healthcare.

According to recent records, the global medical robotics market is currently valued at approximately thirteen billion dollars, with projections to grow even further in the near future. It is essential to conduct further developments within the medical robotics industry, as there is a rise in the United Kingdom alone of six percent, equalling nineteen percent of individuals with a disability since 2013. (House of Commons, 2025) suggests that "from the Department for Work and Pensions' Family Resources Survey indicate that 16.1 million people in the UK had a disability in the 2022/23 financial year. This represents 24% of the total population." This implies the need for further advancements to enhance the quality of life and essential care.

Robot dogs could also help ease the demand for guide dog partnerships and be considered for individuals who suffer reactions to specific allergens, who live in restricted spaces, or are not permitted to have pets.

## 1.2. Objectives

This paper will explore how to answer the research question while incorporating relevant data effectively. Firstly, the study involves running simulations in Webots, Cyberbotics Ltd. (1996) with potential expansion to Gazebo, pedestrian robot Cyberbotics Ltd. (1996), for dynamic obstacle avoidance, and using a BIOLOID robot, Cyberbotics Ltd. (1996) quadruped model equipped with various sensors that can be adapted for further research, including cameras, LiDAR, GPS, an inertial unit, and distance sensors.

The algorithms being tested include Q-learning (Q), Deep Q-networks (DQN), and Proximal Policy Optimization (PPO), all reinforcement learning algorithms.

The simulations will occur in virtual environments with varying complexities, ranging from a simple indoor maze to a dynamically complex scenario. The mazes are custom-made to compare performance and performance metrics across each algorithm (PPO, DQN, and Q). All three will be tested on the same simple map and then in the same dynamic environment. The robot will have the same start and goal nodes, which also helps to control the experiment. The goal is to have a fully functioning autonomous robot dog capable of recognising and avoiding obstacles and following path planning logic. This experiment is to see if any algorithms can outperform the other, and if any of the chosen algorithms can come close to decision-making logic as a well-trained guide dog.

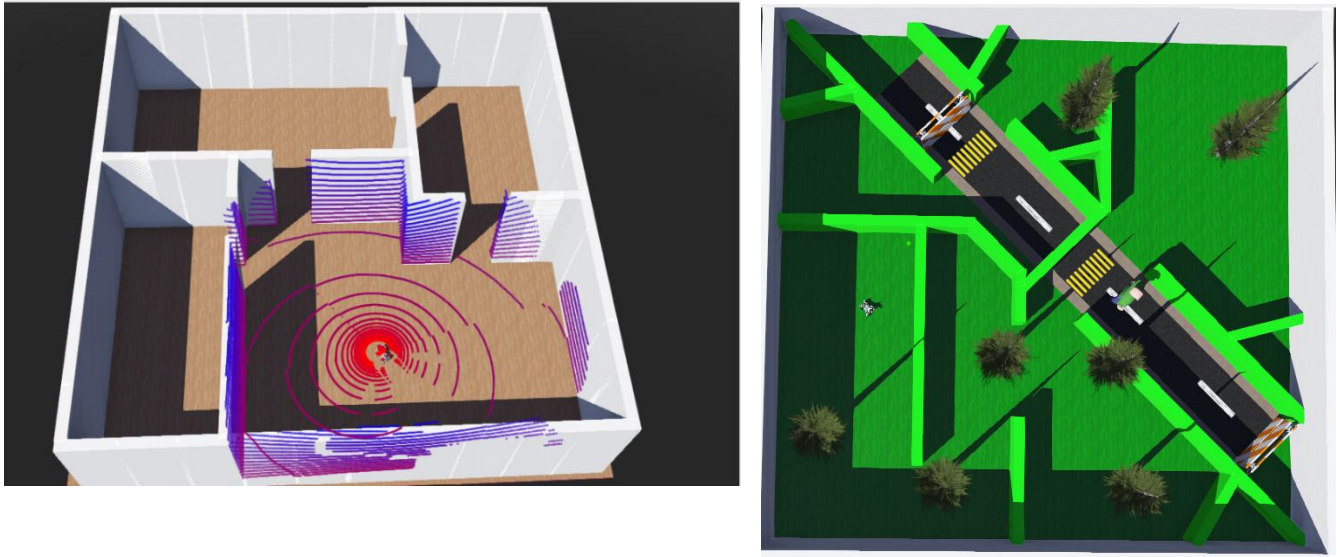

Fig. 1. A simple environment built in Webots (left), for the quadruped to explore, with LiDAR activated. A Dynamic environment (right) for the quadruped to explore

### 1.3. Research Questions and Hypothesis

This paper will evaluate research on whether robotics can help meet the demand for guide dogs and assistance animals. How effective are reinforcement learning algorithms in training a simulated robotic quadruped to navigate complex environments, and how do their capabilities compare to those of trained guide dogs?

This research uses comparable raw data, including sensor data, algorithms, and graphs. This paper will compare reward per episode, the learning rate of each algorithm, the number of steps taken, the average steps to the goal per episode, and the number of collisions. For further information, please refer to 3.3. Results. This is appropriate as the graphs will show the training data to compare the above metrics between PPO, DQN, and Q in the simple and dynamic environments. Further research may be taken, specifically a difference in means, which may be an appropriate statistical analysis. Comparing the average performance metrics between three algorithms, thus assessing the difference in means, is applicable. This involves determining whether any observed difference is statistically significant, which could lead to the appropriate statistical tests (see 3.1 methods for further information).

So far, the hypothesis for this research indicates that there may be an alternative hypothesis, not a null hypothesis. Similar research shows that a quadruped can learn complex environments to a comparable level of intelligence as a guide dog. So, another way to interpret this research question could be that there is a difference in means in evaluating reinforcement learning algorithms for navigation in simulated robotic quadrupeds: A comparative study inspired by guide dog behaviour.

Currently, the Q-learning algorithm is predicted to be the least efficient of the three algorithms, namely Q-learning, DQN, and PPO. Nonetheless, it is too early to have a truly accurate prediction. However, overall, it is believed possible for a quadruped robot to navigate complex environments to the same degree as a guide dog. As previous findings have indicated, this is possible; further research could be valuable when expanding algorithm testing.

### 1.4. Ethical, Legal, and Social Issues

For this specific research, formal ethics approval is not required as there is no human participation or collection of personal data. This research purely depends on evaluating one algorithm against another to see if a robot dog could develop complex behaviour to the same degree as a well-trained guide dog. This paper does not aim to replace guide dogs with robot dogs; it merely aims to experiment with whether a robot dog can achieve the same level of safe engagement as a guide dog, intending to extend to further research within medical robotics, specifically with alert pets.

Regarding legal issues, some potential concerns could revolve around collecting and storing data from the robot's navigational sensor data, as the robot navigates in public spaces.

As this research strives to see quadrupedal robots assist the visually impaired individuals, they should only be seen as complementary rather than replacements. However, some further concerns could derive from affordability, accessibility, and technological inequality.

Ethical decisions would be essential for future world applications, especially when running robotic navigation in real locations and with vulnerable individuals.

### 1.5. Potential Contributions

This paper hopes to contribute to the field of medical robotics, with the intent to further research on integrating medical quadrupeds into patient care. Thorough research into algorithmic comparison and pushing the boundaries of what simulated robots can achieve will enable the medical robotics field to encourage the use of robotic guide dogs where necessary.

Furthermore, future research could be carried out on integration with robotics, not just for partially sighted individuals, but also for medical alert dogs and other quadrupedal uses in the medical robotics field. This paper also allows researchers to critically evaluate different solutions for individuals who need medical pets but cannot own them for various reasons.

The next chapter presents a literature review, followed by the ongoing development, methods, and results captured. Next, the conclusion and future works will follow, where you can expect to find potential further research and implementations. Finally, references will come next, along with an appendix.

## 2. Literature Review

When analysing similar pre-experimental data, ratio data, and research, some existing research papers were more closely linked to what this paper tries to achieve than others. Specifically, some comparisons found that other papers also only researched simulating robots, without using hardware. Some included reinforcement learning, while others sought to research agile navigation. All the documents listed have some aspect in common with this paper. This paper has closely considered all aspects from path planning, collision avoidance, sensors, simulation platform, algorithms, scenarios, hardware deployment, and robot type.

**Literature with useful insights, though not directly related to this paper's experimental procedure.**

A computer simulation program was used to analyse teaching robot navigation in the presence of obstacles. The focus is on path planning and uses three separate techniques: the Potential Field Method, the Vector Field Histogram Plus Method, and the Local Navigation Method. These methods are unique compared to many other papers focusing solely on reinforcement learning; however, they are not directly relevant as the paper focuses on student learning and user-friendly functionality. However, this project still has some potential applications, as a grid-based simulation is being considered for obstacle avoidance for the quadruped 'guide' dog, particularly for Q navigation.

According to (Erin, B., *et al.,* 2010, pp. 567), "In this histogram grid, each cell has a certainty value $c_{i,j}$ which has value one where we are confident there exists part of an obstacle and has value zero where there is no obstacle." Here, the specific technique used was the Vector Field Histogram Plus Method, which allowed the grid-like method to use binary values to determine if an obstacle is present, which will be valuable when implementing this method for Q-learning.

When looking into Sim-to-Real: Learning Agile Locomotion For Quadruped Robots, this paper also utilises a reward system for reinforcement learning and thus offers other considerations, such as using the Proximal Policy Optimisation algorithm in PyBullet.

Although this paper closely focuses on the sim-to-real, its techniques for bridging the reality gap for hardware deployment do not closely follow the same research goals. (Tan, J., *et al.,* 2018, pp. 2) states "In this paper, we choose to use Proximal Policy Optimization (PPO) [5] because it is a stable on-policy method and can be easily parallelized." This method may also be specifically helpful for reinforcement learning for locomotion tasks.

Next, the study: Quadruped Robot Obstacle Negotiation via Reinforcement Learning. This paper heavily focuses on obstacle navigation and reinforcement learning, which coincide well. There are some highly relevant applications of reinforcement learning to quadruped obstacle navigation, and it also demonstrates hierarchical reinforcement learning with a policy search-based low-level controller. However, there is no explicit pathfinding for navigation. (Lee, H., *et al.,* 2006, pp. 1) Express "Our method is based on hierarchical reinforcement learning [19]–[22] using a two-level hierarchical decomposition of the task. Given an obstacle (such as a step) that the robot needs to climb over, the high-level planner selects a sequence of "target foot placement positions" for the robot, one foot at a time." Again, this paper is heavily focused on reinforcement learning, specifically for foot placement, which allows for some interesting insights, although not directly relevant. The hardware of this robot also uses sensors of interest, such as the Inertial Measurement Unit and Servo Encoders.

Considering the paper: Comparison of path-planning methods for robot navigation in simulated agricultural environments. This paper focuses on path-planning logic; however, it only centers on search algorithms: Depth-First Search (DFS), Breadth-First Search (BFS), and the A* algorithm. This paper shares some enlightenment as classical path-planning methods will be implemented in future research, especially as A* could be considered for future path-finding logic. This allows for a hybrid approach combining A* pathfinding with reinforcement learning methods.

This paper above, however, is not directly relevant to quadruped reinforcement learning. (Vásconez, J. P., *et al.,* 2023 pp. 901) It states, "It calculates heuristic functions value at each node in the work area and involves the checking of too many adjacent nodes for finding the smallest cost (least distance travelled, shortest time, etc.)" This clearly explains how the heuristic function works and shows how this algorithm is worth considering when looking into path planning logic.

This paper shows that obstacles have been predefined and solely considers path planning navigation. However, the performance metrics evaluated could be similar. Time and distance travelled are good evaluations of efficiency.

Looking into Self Modeling and Gait Control of Quadruped Robot Using Q-Learning Based Particle Swarm Optimization. This paper does not have too many similarities in research focus; however, looking into more reinforcement learning, namely Q-learning, is important for research diversity. This paper explores the integration of Q-Learning with Particle Swarm Optimization (PSO) to improve the gait control of a quadruped robot. This uses a hybrid style approach where the robot will learn more efficient, balanced walking patterns and gaits. Both approaches show that the robot learns to adapt to its environment over time, but this is where the similarities end. According to (Meerza, S.I.A. and Uzzal, M.M., 2019). "The proposed algorithm is a combination of Q-Learning and Particle Swarm Optimization (PSO). Here the actions are selected using the Q-learning algorithm." This shows that although the Q-learning algorithm is not usually optimal for quadrupedal adaptation, hybrids are still being implemented, and simple tasks are completed to high standards.

**Literature that links directly to this paper's experimental design.**

For similar research, consider synthesizing the optimal gait of a quadruped robot with soft actuators using deep reinforcement learning. According to (Ji, Q., *et al.,* 2002, pp. 7), "Reward function: Our objective is to encourage the robot to walk straight for a longer period while keeping a steady gait…" displays that a reward system was considered over a punishment system when the robot can keep a steady gait. Although the Bioloid robot has a controller for specific gait functions, the quadruped will use reinforcement learning for specific path planning logic. A reward system would also work in the context of this research paper. However, a punishment system will be used alongside a reward system, making for different and more interesting outcomes. A more intense ratio of punishment and reward will be used to influence correct decision-making by the robot.

Other aspects inspired by this paper included path planning using RL combined with LiDAR, camera-based sensors, and ultrasonic sensors. The scenarios also extend to outdoor environments and complex obstacles.

Whilst researching ANYmal Parkour: Learning Agile Navigation for Quadrupedal Robots. Again, the techniques used here are the Hierarchical Navigation Framework, and the navigation policy is learned through reinforcement learning. Here, (Hoeller, D., *et al.,* 2024), also uses LiDAR for experimentation, "We decompose the problem into three components: the perception module receives the point cloud measurements to estimate the scene's layout and produces a latent tensor and a map." A lot of insight will be taken from this paper as it is highly relevant due to its fully learned approach to quadrupedal navigation. This paper also demonstrates reinforcement learning-based pathfinding and locomotion in real-world settings and includes both perception-driven obstacle avoidance and agile locomotion strategies. This could strongly link to guide-dog-like navigation problems.

LiDAR and depth cameras are also used for obstacle reconstruction, and the ANYmal also utilizes an inertial measurement unit (IMU) and joint encoders.

The scenarios for testing this robot in a parkour-like environment are more complex, with randomised terrain including gaps, steps, various heights, and narrow platforms.

So, a potential application could be implementing a perception module for 3D scene reconstruction, which could improve guide-dog applications. This research also considers integrating learned locomotion policies rather than only classical path-planning algorithms.

Learning Forward Dynamics Model and Informed Trajectory Sampler for Safe Quadruped Navigation is a closer match in terms of research. Again, the focus is on quadrupedal navigation using reinforcement learning. This addresses real-time trajectory optimization and obstacle avoidance, which are critical for a guide-dog-like robotic system.

It also incorporates LiDAR-based perception to navigate complex environments dynamically and combines deep learning with classical path planning methods for robust navigation.

Potential integrations from this research paper could include incorporating the Forward Dynamics Model (FDM) to improve collision avoidance in guide-dog applications. The Informed Trajectory Sampler (ITS) could be trained on real-world guide dog movements. Also, the sampling-based MPC approach ensures safe, dynamic adaptation to obstacles, crucial for assisting visually impaired users.

According to (Kim, Y., *et al.,* 2022, pp. 1), "In this work, we propose a learning-based fully autonomous navigation framework composed of three innovative elements: a learned forward dynamics model (FDM), an online samplingbased model-predictive controller, and an informed trajectory sampler (ITS). Using our framework, a quadruped robot can autonomously navigate in various complex environments without a collision and generate a smoother command plan compared to the baseline method." These three navigation frameworks allow the robot to navigate fully autonomously, which makes the FDM and ITS models very attractive for research involving smooth path finding and collision avoidance.

Another paper that closely matches this research is Learning robust perceptive locomotion for quadrupedal robots in the wild. Similarities include its provision of a strong reinforcement learning-based framework for quadrupedal locomotion. Here, the reinforcement learning policy selects the best actions dynamically through Adaptive Curriculum Learning. LiDAR, Depth Cameras, IMU, and Joint Sensors directly apply to this paper's guide-dog concept. It follows a structured approach to terrain adaptation, similar to the three levels of difficulty considered.

Expressing (Miki, T., *et al.,* 2022, pp. 17) "We used our custom implementation of PPO [59] to train the teacher policy [70]. Observations are normalized using running mean and standard deviation before giving them to the policy network." This shows that PPO has been used again and is a popular choice for reinforcement learning frameworks.

The following paper presents a hierarchical path planning method for quadruped robots that relies on dynamic 3D point clouds to improve navigation in complex environments. Dynamic 3D Point-Cloud-Driven Autonomous Hierarchical Path Planning for Quadruped Robots integrates Global Path Planning using Particle Swarm Optimization (PSO) and Artificial Potential Field (APF), and also Local Path Planning using an Improved Dynamic Window Approach (DWA).

This paper is very relevant regarding similar research, particularly in path planning and collision avoidance; however, it does not have anything in common with robotic guide dogs' simulation research. (Zhang Q., *et al.,* 2024, pp. 259) share great insight into the difficulties in overcoming complexities within path-planning. "In summary, traditional path planning algorithms for quadruped robots face the following issues: (1) The environment map, composed of idealized regular geometry, is used as the input of the algorithm, which cannot effectively plan the path in real complex environments…" This shows that careful planning and considerations are required to ensure accurate results in this type of experimental design.

None of these algorithms are reinforcement learning based; even the sensors are completely different, where stereo cameras are used here. One similarity here could be that APF is also a heuristic-based algorithm like A*, which will be potentially used for further research. The 3D Point-Cloud-Driven Lidar will also be potentially helpful in this paper.

Learning the Quadruped Robot by Reinforcement Learning (RL). This paper investigates using Deep Deterministic Policy Gradient (DDPG) Reinforcement Learning to control a quadruped robot and compares it against a PID controller-based approach.

It is simulated using the SimScape-Multibody toolbox in MATLAB. Again, the use of RL is included in this paper. However, it's different because this is the only paper integrating a PID controller.

It also purely focuses on locomotion, not obstacle navigation and planning complexity. According to (Issa, A.A., and Aldair A.A. 2022, pp. 121), "Because we want the robot to walk in a straight line forward, the first reward is added forward velocity (Vx) in the x-direction (which means that the robot moving in forward in a straight line). The second and third rewards penalize the y and z dimension displacement, ensuring the robot does not deviate too far from the line or fall." It focuses on a multiple reward system where the reward function/negative reinforcement forces the robot to improve. It is an interesting concept for using both positive and negative reinforcement techniques, which will also potentially be tested for this guide dog research.

Researching into Gait Learning for Hexapod Robot Facing Rough Terrain Based on Dueling-DQN Algorithm has many similarities in terms of research. The paper proposes a gait learning method for hexapod robots that must navigate hazardous and uneven terrains, such as slopes, terraces, and ravines.

It focuses on how well the Dueling Deep Q-Network algorithms allow the hexapod robots to autonomously control their foot placements based on different terrains. LiDAR is also used for environmental perception, and a variant of DQN is also applied here. Both this paper and research into robot 'guide dogs' have potential for real-world applications, and if there are any miscalculations, both could have devastating effects.

However, research into robotic guide dogs is simulation-based only, unlike this paper, which also has physical deployment. As seen by (Chen, L., *et al.*, 2024), "Utilizing robots instead of human workers for these tasks can mitigate these risks [2]. Therefore, selecting suitable robots and devising methods for their movement in rugged terrain are essential for handling dangerous goods in explosive environments."

This next report is pioneering work; it displays one of the earliest explorations into quadruped locomotion and captures foundational concepts that have influenced modern research in legged robotics. Design of a Quadruped Walking Vehicle.

This research focuses on the control strategies, mechanical design, and the challenges faced in achieving stable early locomotion. The goal was to be able to create a physical machine that could assist in certain environments that are specifically inaccessible to wheeled vehicles. Although the technology is now outdated, there are good insights into early mechanical design and control strategies.

Again, this paper looks into stable walking and indicates many complexities that surround this area. This paper displays that there are still many similarities when researching efficient mechanical design for quadruped robots, and explains that the persistent complexities surrounding mechanical design remain consistent even across decades. (Tachi, S., *et al.,* 1985) displays early considerations into real-life obstacle navigation. "In order to detect these obstacles, an ultrasonic sensor, which can determine not only the distance from the obstacle but also its direction by the traveling time measurement of ultrasound was developed."

However, as expected, there are many drawbacks to this research, possibly due to the time and limited technologies of this era. There is limited experimental data and a lack of visual aids, which in turn make it challenging to assess its effectiveness in this area.

## 3. Ongoing Development

### 3.1. Methods

This section of the paper focuses on the approach and techniques taken in this area of study, which were influenced by the previous research papers discussed above. Firstly, the techniques discussed will be the different types of learning methods. As stated previously, a range of other learning methods helps analyse which learning method is most effective for path planning.

Reinforcement learning seems inclusive when implementing path-finding logic; however, perhaps one algorithm is more effective than the rest. This paper aims to discover this and determine if a simulated quadruped really could 'act', 'think', and 'react' to complex situations as a guide dog can.

This paper would benefit from assessing Q-learning and its effectiveness compared to different learning methods. Q-learning aims to learn an optimal policy for an agent by iteratively updating a Q-table, where rows represent the states *(s)* that the agent can be in, columns represent possible actions *(a)* the agent can take, and each cell stores a Q-value, *Q(s,a,)*, which estimates the future reward of taking an action in a state. The agent explores the environment using the Bellman equation:

$$Q(s,a) \leftarrow Q(s,a) + \alpha \left( R + \gamma \max_{a'} Q(s',a') - Q(s,a) \right) \quad (1)$$

Equation (1) shows the mathematics behind the Bellman equation.
Where:
- *Q(s,a)* is the current Q-value for state *s* and action *a*
- α is the learning rate, which determines how much the Q-value is updated each step
- *R* is the immediate reward received after taking action *a* in state *s*
- γ is the discount factor, which controls how much future rewards are considered
- $\max_{a'} Q(s',a')$ is the highest Q-value for the next state, *s'*, representing the best possible future reward
- *s'* is the next state after taking action *a*.

Firstly, initialise the Q-table so all values are set to zero. Next, repeat for each episode. Start at the initial state and choose an action using an exploration strategy. Once the action is executed, the agent moves to the next state, and the reward is received whilst updating the Q-value. This process will be repeated until the goal is reached, and after many episodes, the Q-table converges, allowing the agent to learn the optimal policy.

The advantage of using Q-learning is that it is model-free, so no prior knowledge of the environment is needed; this is helpful as the agent can explore and learn the environment without predefined rules. Q-learning is simpler to implement than Deep Q-Networks as it does not require neural networks. Q-learning guarantees convergence. Suppose the learning rate α is tuned correctly. In that case, Q-learning is mathematically guaranteed to converge to the optimal policy in finite, discrete state spaces, which makes it reliable for small-scale problems like grid-world navigation.

Motivation for this paper's Q-learning algorithm has arisen from several sources that were further adapted for DQN. Ronan Murphy's GitHub (Murphy, 2020), Gambino's GitHub (Gambino, 2017) and MinRL (10-OASIS-01, 2025). The MinRL project offers a Q-learning agent for tabular Q-learning, which is highly similar in structure to this paper's grid navigation Q-learning agent. It maintains a Q-table for state-action values and employs an ε-greedy policy with decay for exploration. Each update uses the standard Q-learning rule and the 'update_q_value' method. MinRL's agent decays its exploration rate over time and iterates over episodes to learn optimal navigation. The code is organized in a precise class with methods for choosing actions and updating the Q-table, making it functionally and structurally equivalent to this research paper's implementation of tabular Q-learning solution.

However, one setback to consider is that Q-learning has poor scalability in large state spaces, as values for every state-action pair are stored in a table; larger spaces could mean larger tables and thus be too hard to manage. Like other reinforcement learning techniques, it could be slow to learn, meaning a greater need for many episodes.

This paper also evaluates the effectiveness of Deep Q-Networks (DQN), which can be seen as an extension of traditional Q-Learning. While Q-Learning uses a table to store the Q-values for each state-action pair, DQN replaces this table with a neural network that approximates the Q-function. This makes it much more scalable for environments with large or continuous state spaces, where Q-tables become impractical and could get out of proportion. DQN works with Q-learning, implementing deep neural networks to learn optimal policies.

The agent tries to learn an optimal policy by estimating Q(s, a), the expected future reward of taking an action, a, in state, s. Still, instead of storing all values, it learns them through function approximation. Now, mathematics has become updated:

$$\text{LOSS} = \left( R + \gamma \max_{a'} Q_{\text{target}}(s', a') - Q(s, a) \right)^2 \quad (2)$$

Equation (2) Shows the continued mathematics behind the Bellman equation.
Where:
- Q(s, a) is the predicted Q-value from the current network
- $Q_{\text{target}}(s', a')$ is the target Q-value from a separate target network
- α is the learning rate, which determines how much the Q-value is updated each step
- R is the immediate reward received after taking action *a* in state *s*
- γ is the discount factor, which controls how much future rewards are considered
- $\max_{a'} Q(s', a')$ is the highest Q-value for the next state, *s′*, representing the best possible future reward
- *s′* is the next state after taking action, *a.*
- LOSS is used to train the network using backpropagation.

To stabilise training, DQN introduces two essential mechanisms:

Experience Replay, which stores past transitions in a memory buffer and samples small batches randomly to break correlation between sequential data.

Target Networks in which a copy of the Q-network is updated less frequently to provide more stable targets for learning.

DQN works by starting with a randomly initialised neural network. During each episode, the agent selects actions (this paper has used an ε-greedy exploration strategy) that store the experience in the replay buffer, sample batches from it, and update the Q-network accordingly. Over time, the model learns to approximate optimal Q-values and policies.

Again, MinRL has influenced how DQN is implemented in this paper (10-OASIS-01, 2025). MinRL is an educational project focusing on fundamental RL algorithms in a grid-based environment. It includes a full DQN agent implementation in PyTorch that matches this paper's DQN algorithm design. Key similarities include a neural Q-network function approximator, an experience replay buffer for sampling past transitions, and a separate target network to stabilize training. The agent uses an ε-greedy exploration strategy with decaying epsilon, as this version does. The training logic (batch updates of Q-values using the Bellman equation, periodic target network updates, and using a Huber/MSE loss).

Two specific networks are experimented with here: an online and a target Q-network. Both consist of three hidden layers and 272,644 total parameters, weights, and biases, with three hidden layers.

One advantage of DQN is that it handles large/continuous state spaces much better than Q-learning due to its use of function approximation. It retains Q-learning's benefits, specifically off-policy and model-free, while significantly expanding its capability to complex environments. Also, it can solve problems that Q-learning cannot solve due to memory constraints.

DQN also has disadvantages, such as requiring more tuning. Neural networks introduce more hyperparameters and instability, such as overfitting and divergence. Also, because of the additional memory and backpropagation, DQN is slower to train. Finally, there is no convergence guarantee if the function approximation or the exploration strategy is poor.

The third algorithm assessed in this paper is Proximal Policy Optimization (PPO), a policy gradient method in the class of on-policy reinforcement learning algorithms. Unlike Q-Learning and DQN, which are value-based and learn what action to take based on estimating value functions, PPO is a policy-based method that directly learns the optimal action probabilities using a neural network policy. PPO works especially well when working with both discrete and continuous action spaces.

PPO aims to update the policy in small, stable steps, avoiding large jumps that might destabilize training. The PPO objective function penalizes significant changes in policy with a clipping function, which ensures the new policy does not deviate too far from the old one during an update. PPO uses Actor-Critic architecture, policy gradient method, and clipped surrogate objective. The clipped objective looks like this:

$$L(\theta) = \mathbb{E}[\min(r(\theta)\hat{A},\ \text{clip}(r(\theta), 1 - \varepsilon, 1 + \varepsilon)\hat{A})] \quad (3)$$

Equation (3) shows the mathematics behind PPO's clipped surrogate objective.
Where:
- $L(\theta)$ is the loss function
- $r(\theta)$ is the probability ratio, which stands for

$$\frac{\pi_\theta(a \mid s)}{\pi_{\theta_{old}}(a \mid s)} \quad (4)$$

- Where $\pi_\theta$ are the parameters of the updated policy
- $\pi_{\theta_{old}}$ are the parameters of the previous policy
- $(a \mid s)$ is the probability of taking action *a* when in state *s*
- $\mathbb{E}$ is for the expected value, the average over many samples
- And $\hat{A}$ stands for the estimated advantage
- $\varepsilon$ is the clip threshold.

Equation (4) is important to understand, as the probability ratio explains how likely the new policy is to take action, *a,* in state, *s*, compared to how likely the old policy was to take that same action.

Now training proceeds with:
1. Data collection collects and stores data as the agent interacts with the environment.
2. Next, the function estimation allows the Critic to evaluate the current policy.
3. The advantage function is now calculated, which allows for estimating if an action was better than expected or worse.
4. The policy is now updated. The Actor is updated based on the advantage and clipped surrogate objective functions.
5. Now repeat.

PPO balances performance and simplicity, offering a solid way to improve policies without the complexity of trust-region methods. PPO uses a shared layer, Actor Head, and a Critic Head, allowing one unified network with two heads: action probabilities and state values.

Inspiration has been taken from geekyutao's PPO, (geekyutao, 2018), Shahbazi's Webots repo, (Shahbazi, 2023), CleanRL repository (vwxyzjn, 2025). Also, from 10-OASIS-01, MinRL (10-OASIS-01, 2025), it features an actor-critic architecture with separate policy (actor) and value (critic) networks, using a clipped surrogate objective and Generalized Advantage Estimation (GAE) for advantage calculation. The training loop (mini-batch updates, entropy regularization, etc.) is similar to this paper's algorithm implementation. - CleanRL's code collects trajectories, computes advantages, and applies the PPO clipping rule the same way this paper does; it also includes handling of entropy and value loss. PPO has 274,437 total parameters between the shared layer, Actor Head, and Critic Head, with three hidden layers, one is a shared input.

The training approach for testing multiple path planning algorithms in this research will be simulation-based testing. The study will be conducted by running the simulation multiple times with different algorithms to record their performance based on key metrics. Each algorithm will be evaluated in a controlled, repeatable environment to compare efficacy, success rates, and collision avoidance.

Firstly, a simulated environment will be set up in Webots with a range of simple to complex scenarios, with both static and dynamic obstacles. Each path planning algorithm, PPO, DQN, and Q, will be implemented and run multiple episodes, within one long simulation for each environment. Here, the data will be collected and analysed. Specific tests include reward per episode, learning rate, steps per episode, average steps to the goal, and the number of collisions recorded. This allows each algorithm to be directly collated and evaluated objectively through measurable metrics. Webots has been selected as a good platform for simulation-only research. It is also incredibly user-friendly without needing a steep learning curve to operate. Although Webots is not currently the industry standard for robotic simulation, it makes for interesting research, as few other papers have used Webots for research purposes.

What should this research paper's experimental design include? Starting with the variables, the independent variable will be the three learning algorithms, PPO, DQN, and Q. The dependent variables will be the measurements of each outcome, success rate, collision count, learning speed, reward per episode, and others. For more information, see 3.3 Results. The controlled variables will be that the same quadruped, along with the same sensor type and amount, will be used throughout each test. Also, the same number of iterations will be used for each algorithm in each scenario. Each algorithm will also be tested in the same simple to dynamic environments.

One possible statistical analysis method when comparing results could be the difference in means, as comparing multiple algorithms to see if differences are statistically significant would be a good fit. Furthermore, as this method depends on the dependent variable either being statistically normally distributed or not being normally distributed, and depending on how many algorithms are tested overall, some statistical analyses could include the t-test, the Pairwise t-test, or the Wilcoxon test. Therefore, the visual analysis will likely consist of a boxplot and histogram. This statistical analysis is interesting to evaluate and should be reinvestigated again with future research.

### 3.2. Development

This section provides a brief overview of what to expect, including the datasets implemented so far, details of the setup required, interfaces used, integration, and, ultimately, challenges in deployment.

Firstly, as previously mentioned, the simulation will be run over Webots' platform, and so the algorithms will be tested on Cyberbotics' own robot, quadruped Bioloid, and on Cyberbotics Ltd.'s (1996) controller. The datatype the robot is being trained on is GPS coordinates, camera detection, distance sensory data, data collected from the Inertial Unit, and LiDAR scans.

The experimental setup consists of Webots' own API, but can be further experimented with by implementing ROS and ROS-dependent packages such as roscpp and rospy along with PyBullet. The code is fully programmed in Python, and the simulated environments are tailor-made with Webots' own PROTO Nodes.

Firstly, sensors being explicitly tested in this paper are the inertial unit (IMU), touch sensors, camera, and navigation, which are expressly tested using LiDAR and GPS. As Q-learning is ready for testing, which works with discrete spaces, the real-world coordinate system has been converted into a twenty-by-twenty grid for a ten-by-ten-meter space, allowing each cell to be zero point five meters. It helps dictate future actions when deciding on cell collision. The cell is either empty, full, or partially full, depending on whether an obstacle is found in this position. This is executed before any path-finding logic is implemented. Now, the quadruped can choose from four discrete actions: forward, turn left, turn right, and wait. Each action will have consequences, namely:

- Collisions will be penalised at -100
- Penalty for falling will be -50
- Scaling punishment from -10 to -1. Greater punishment is given towards the starting point, and less is given with progression towards the goal node.
- A reward of 100 is delivered on reaching the goal.

The algorithms allow for one long simulation consisting of a capped 20,000 episodes. The robot will be repositioned at the starting coordinates for every successful completed run and every fall.

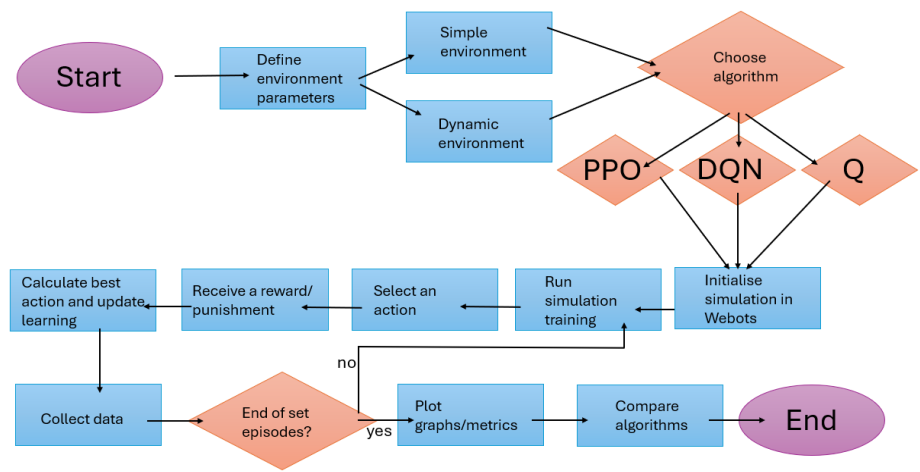

Fig. 2. Shows a flowchart of the experimental procedure for visual ease.

Q-learning, DQN, and PPO are three reinforcement learning algorithms that are being considered for testing on the quadruped guide dog. PPO seems to have a lot of popularity within similar research and has proven to be quite effective when comparing path planning, obstacle avoidance, and foot placement. Although DQN also has reasonable popularity in similar research, Q-learning has very little experimental data in this type of research, and thus, it would bring an interesting perspective. However, as the quadruped already has both a low- and high-level of control, reinforcement learning will help with collision avoidance. This paper's research will focus exclusively on path planning logic.

Initially, the focus of this paper was to compare different algorithms; the focus was to compare the efficiency of reinforcement learning algorithms, genetic algorithms, and search algorithms. The tested algorithms included a Hybrid A* Search, Canonical Genetic Algorithm (CGA), Q-learning, and Deep Q-Networks (DQN). A range of search, genetic, and reinforcement learning algorithms, respectively. This paper has changed focus through the course of testing; it made sense to compare how well reinforcement algorithms worked, and thus, this avoided needing specific obstacle avoidance algorithms. The robot was initially to be trained in an entropy-based occupancy grid algorithm before any path-finding logic was implemented.

Development challenges included broken ROS packages preventing correct setup, loss of robot interaction, and sensory calibration issues, particularly regarding LiDAR connectivity drops. Many challenges occurred with the supervisor node and function, allowing the robot to reset its position, which helped when the robot continued to fall over. Overall, the smaller challenges consisted of overcoming many debugging issues.

```python
def sendToSupervisor(condition: str, value: bool): #fallen, goalReached, timeout):
    #Send JSON data to supervisor for failure / reset conditions
    #TODO: Input datatypes?
    jsonData = {
                "fall": (condition == "fallen" and value == True),
                "goal": (condition == "goal" and value == True),
                "time": (condition == "timeout" and value == True)
    }
    socketOUT.send_string(json.dumps(jsonData))
```

Fig. 3. Shows a snippet of code in which the main controller talks to the supervisor controller asking to reset the quadruped after a fall.

### 3.3    Results

When evaluating the results, this paper will compare all algorithms' performances across simple and complex environments. Analysis will include reward per episode, learning rates, steps per episode, average steps to the goal, and finally, number of collisions. Please note that, because of time and computational power, I have tried to run for 20,000 episodes but only ended up with roughly 15,000 episodes, as this research is computationally expensive. With further power and more time, this experiment would benefit from at least 50,000 episodes.

Graphs will be compared side by side to clearly show the differences between the simple environment and the complex, dynamic environment.

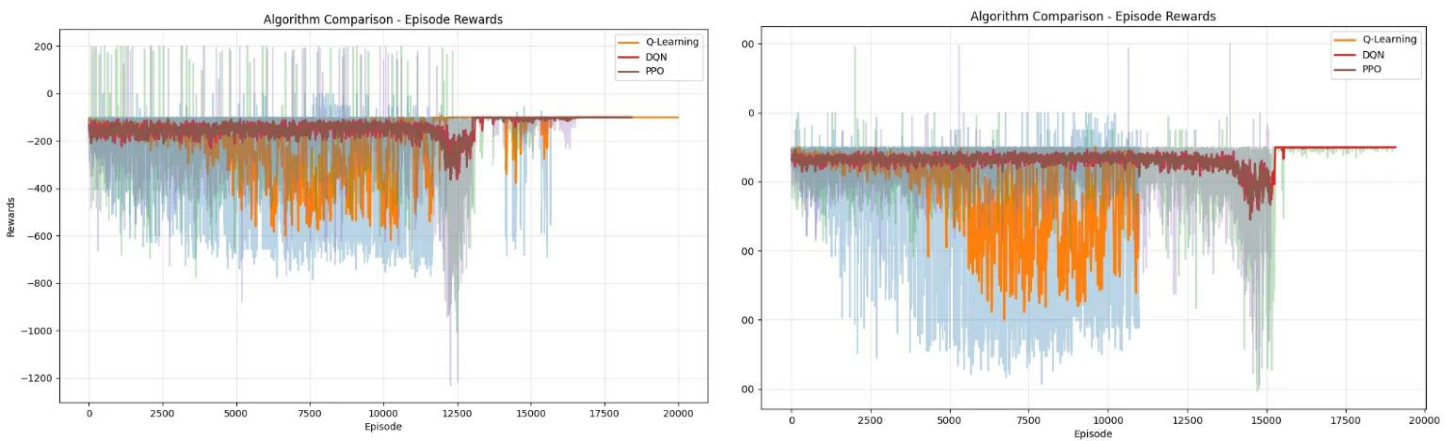

Fig. 4. Graphs to show reward per episode with smoothing for PPO, DQN, and Q. Results for the simple environment on the left and the dynamic environment on the right.

As you can see from Figure 7. PPO performs well; it is not as chaotic as Q-Learning or as stagnant as DQN. All algorithms seem to perform well within the simple environment, but there are significant differences in the dynamic environment. Q-learning does not seem to reach the goal in the dynamic environment at all. There is some improvement, as PPO is the only algorithm consistently reaching the goal. The failures and penalties seem less severe compared to the other two. However, PPO still has a lot of learning to do, as it appears to have emergent progress. Also, the light purple, blue, and green lines show the individual reward/punishment per episode for PPO, Q, and DQN, respectively. Thus, the brown line indicates PPO smoothed performance.

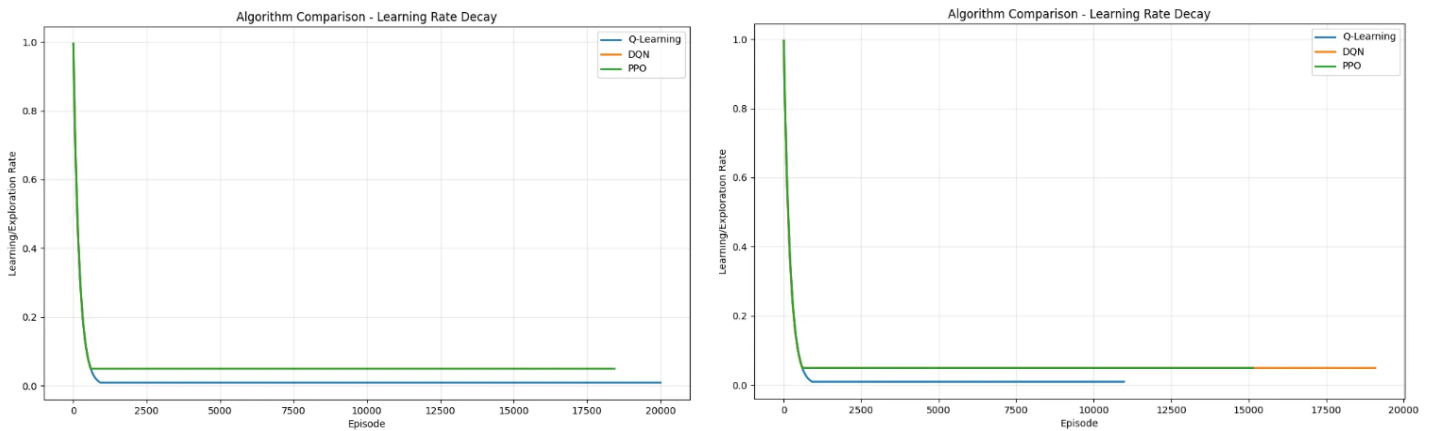

Fig. 5. A graph to show the learning rate between PPO, DQN, and Q. Metrics given for the simple environment on the left, and dynamic on the right.

PPO and DQN's curves show smooth, consistent decay. Unlike Q-Learning, it does not flatline too early. This shows that they both continue to explore long into training, indicating that they haven't stopped learning. Q-Learning drops quickly, potentially leading to premature convergence.

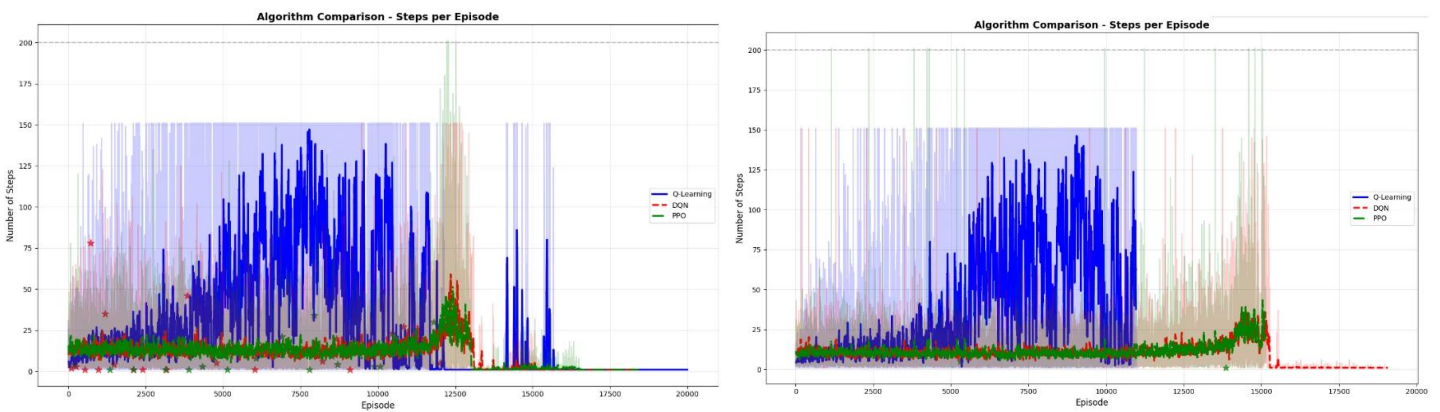

Fig. 6. Graph showing the number of steps per episode between PPO, DQN, and Q. The Goal reached is represented by stars. Performance results from a simple environment on the left and a dynamic environment on the right.

This graph shows that PPO and DQN stay more consistent with gradual improvement. PPO shows that it has taken fewer steps overall and has stayed consistent. All algorithms could easily perform well in a simple environment with multiple goals reached. However, Q is still inconsistent, and even in the simple environment, it gets outperformed.

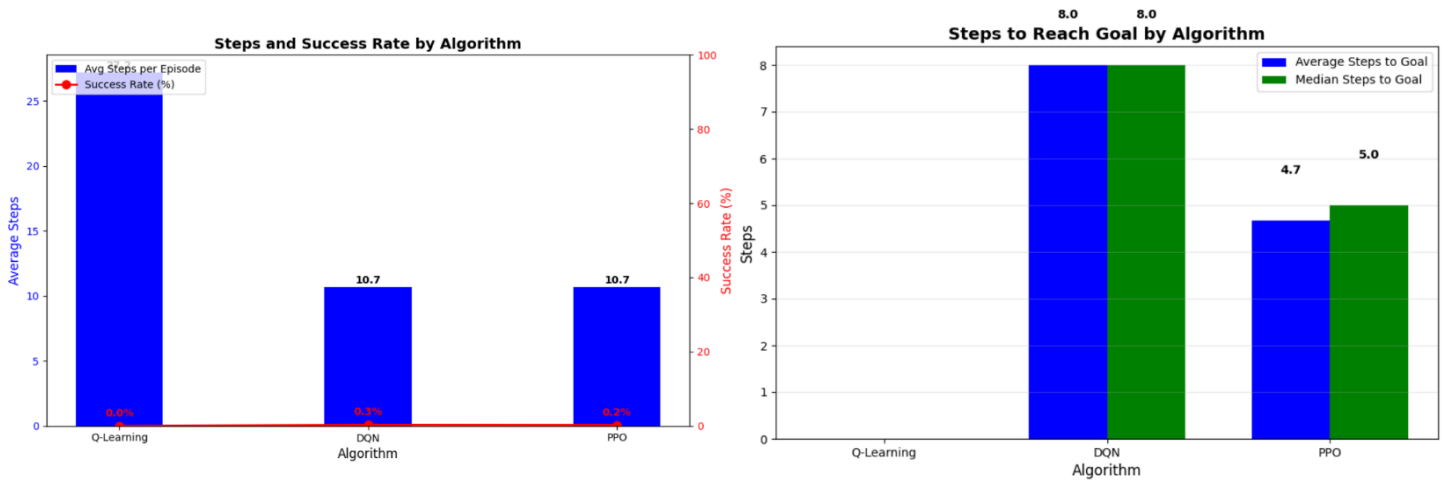

Fig. 7. A graph to show the average and median steps taken to the goal with success rate per episode for PPO, DQN, and Q. Left shows metrics for the simple environment, right for the dynamic environment.

The number of steps to reach the goal per algorithm here shows that PPO has easily outperformed DQN and Q. Q was never even able to reach the goal in the dynamic environment compared to PPO and DQN. The metrics for PPO are almost 50% less than those of DQN. This graph shows strong evidence that PPO is learning effectively. The goal is being reached more directly, with less wandering.

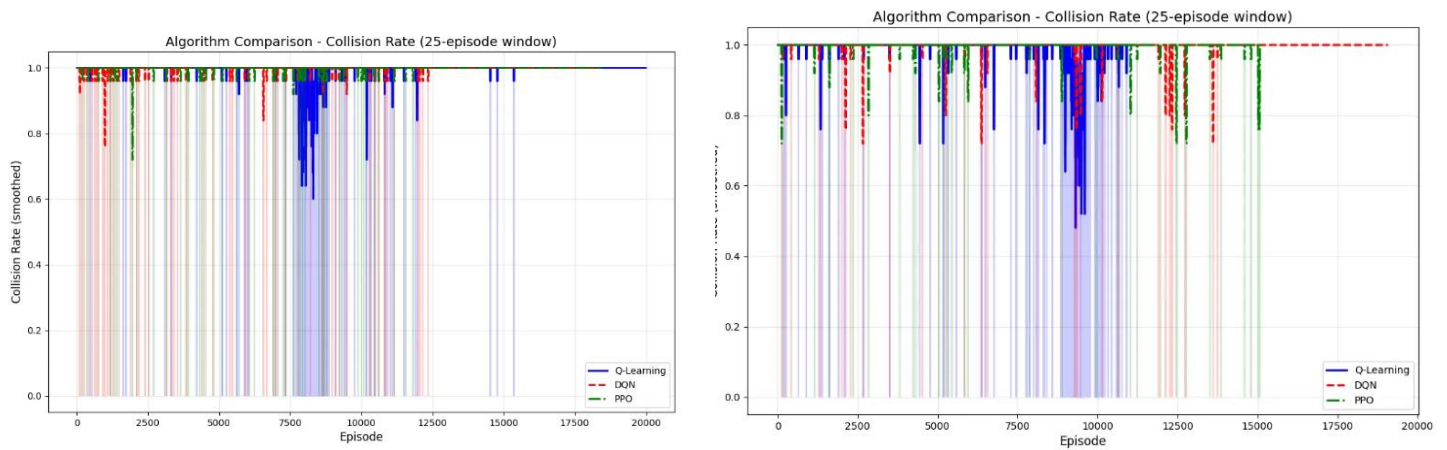

Fig. 8. Shows the number of collisions between PPO, DQN, and Q with smoothing. The simple environment results are on the left, and the dynamic environment results are on the right.

Both DQN and PPO have considerably fewer collisions than Q; however, PPO still consistently collides less than both other algorithms. PPO starts with lots of collisions, and although it's bumpy, it appears to gradually reduce collision rates over time, with more frequent dips below 1.0 later in training. Q-Learning and DQN are bouncing wildly and have less stability. PPO seems to be exploring more cautiously and learning more gradually.

In conclusion, PPO was expected to be the best-performing algorithm, and the metrics (figures 7-11) do indeed show this. This was anticipated through the above research; PPO was an especially popular algorithm to be tested alongside DQN because of its deeper layers and extensive parameters compared to Q-learning, which has no neural network.

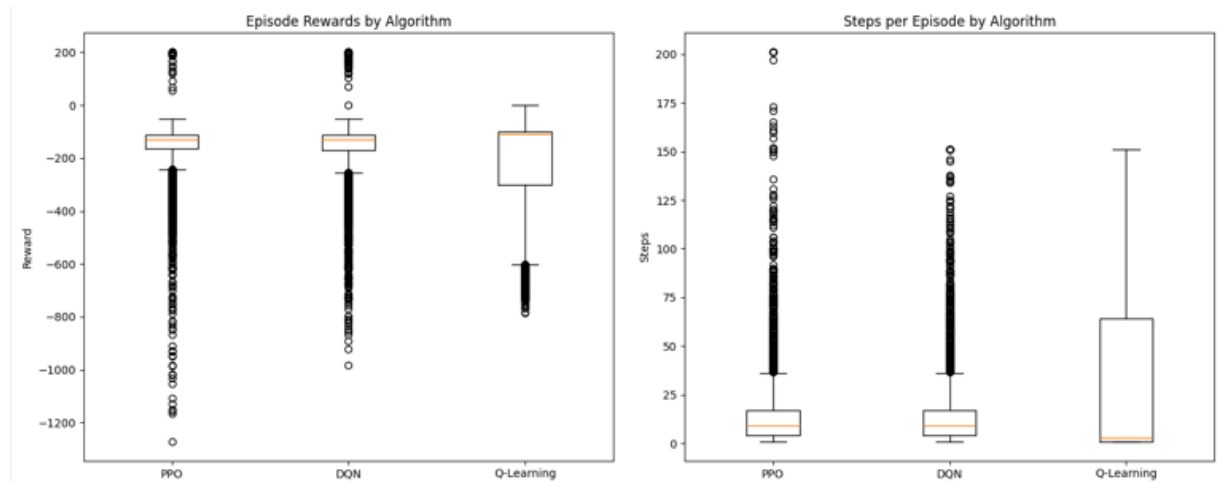

Fig. 9. Shows the statistical analysis of the difference in means of average reward per episode by each algorithm and the difference in means of average steps per episode by each algorithm in a simple environment.

**Simple environment**

When evaluating the simple environment, the average reward per episode by each algorithm is PPO shows the best performance with the highest median reward (around -125) and the most consistent results. DQN follows closely with a slightly lower median reward (around -130) but performs almost equally as well as PPO overall. However, Q-learning performs significantly worse with a much lower median reward (around -300). All algorithms show considerable variation in rewards, but Q-Learning has the most extreme negative outliers reaching down to -1200.

Now, considering steps per episode by algorithm:
PPO and DQN show very similar performance with median steps around 10-15 per episode, and Q-learning takes significantly more steps per episode, with a median around 65. PPO and DQN appear more efficient at completing episodes than Q. Nonetheless, both PPO and DQN have many outliers showing episodes with 150-200 steps

Comparing the success rates by algorithm:
All three algorithms show very low success rates (close to 0). This was expected because training was capped at only 15,000 episodes and does not show the actual efficiency of both PPO and DQN in well-trained dynamic situations. PPO has a marginally higher success rate, but it's barely negligible, and DQN also shows a near-zero success rate. Q-learning appears to have zero success, which was expected because in Fig. 7. (Right), we can see PPO does not reach the goal in the dynamic environment.

In conclusion, PPO performs best overall with better rewards and more efficient movement. DQN is a close second, performing similarly to PPO in most metrics, which makes DQN an equally good candidate for further experimentation. Q-learning performs significantly worse in all metrics.

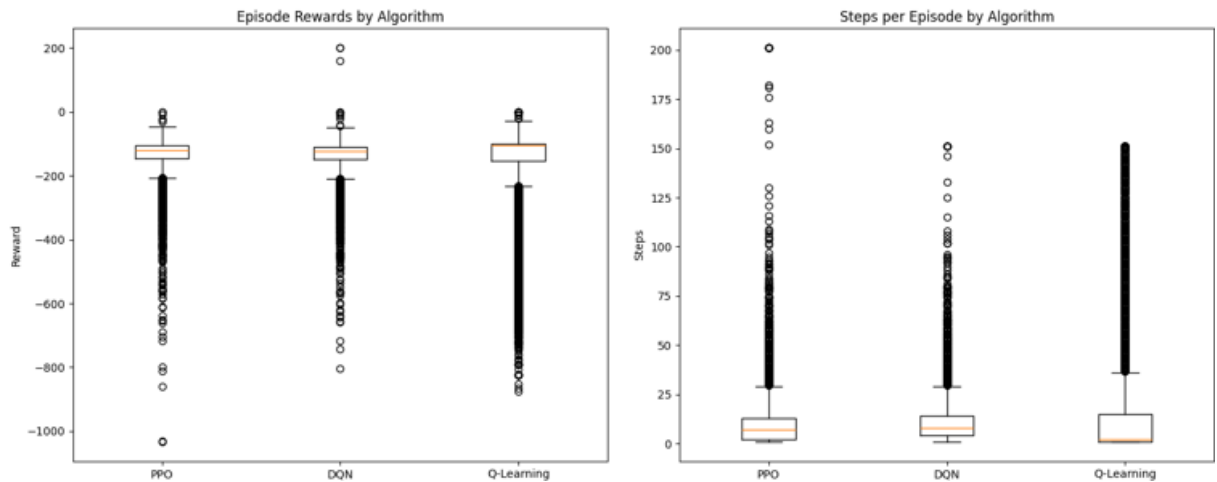

Fig. 10. Shows the statistical analysis of the difference in means of average reward per episode by each algorithm and the difference in means of average steps per episode by each algorithm in a dynamic environment.

**Dynamic environment**

When evaluating the simple environment, the average reward per episode by each algorithm, all three algorithms (PPO, DQN, Q-Learning) show very similar performance with median rewards around -120. The reward distributions are consistent across all algorithms; all algorithms have outliers reaching around -800 to -1000. However, PPO has some positive outliers reaching up to around 200, suggesting occasional successful navigation to the goal.

Now considering steps per episode by algorithm:
PPO and DQN show similar performance with median steps around 10-15 per episode, and Q-learning shows slightly higher median steps (around 15-20) per episode, including unsuccessful episodes. All algorithms have significant outliers reaching 150-200 steps and Q-learning has more extreme outliers (up to 150+ steps)

In conclusion, all algorithms struggled to really perform. Reward distributions are more similar across algorithms in the dynamic map, and the dynamic environment appears more difficult overall. However, interestingly, episode counts vary more as DQN has fewer episodes than PPO/Q-learning.

To summarize, the simple environment allowed PPO to outperform all other algorithms consistently, but DQN is still a good competitor. This was expected with capped episodes. Simple environments should be able to give good metrics compared to dynamic environments with limited episodes, as less exploration is needed and no dynamic obstacles. Although 15,000 episodes were not enough, even for a simple map, it shows that algorithms were able to give solid data and gives an insight on what could be achieved with further time and computational power invested. However, even though PPO didn't reach the goal so often in the dynamic map, it was still able to consistently reach the goal more than DQN and Q. With greater training, I do believe PPO could perform even better in the simple environment, and significantly better in the dynamic environment.

# 4. Conclusions and future work

## 4.1 Conclusions

This research compares different path-planning algorithms (Proximal Policy Optimization, Deep Q-Network, and Q-learning) for quadruped robot navigation in simulated environments. So far, significant progress has been made in simulating capped amounts of episodes. Testing has focused on ensuring accurate sensor readings and reliable environmental perception, which are critical for evaluating navigation performance, including performance metrics on PPO, DQN, and Q in varied environments. In conclusion, PPO did outperform both DQN and Q, which was expected. Ultimately, the goal is to discover if a simulated quadruped robot dog can meet the exact expectations of a guide dog when navigating complex environments. The Q-learning algorithm was predicted to be the most inefficient of the three algorithms. Nonetheless, Q still performed well in simple environments overall.

Considering all the research and metrics supplied, this provides evidence that PPO and DQN can be a foundation for real-world robotic guide dog training. Could the system be integrated into real-world robotic dogs? Yes! However, as has been explored, it is computationally expensive, and training must be run extensively before being commercially distributed.

## 4.2 Future work

Further research will be undertaken, and with the generosity of Dr Diego Resende Faria, who has kindly stated will donate resources to help publish this paper in a Q1 Journal.

Potential areas for improvement consist of more iterations and episodes against which the algorithm can be tested, better sensor collaboration, and exploring additional algorithms and environments. Only three algorithms were tested and compared, as time is a constraint, and as is limiting computational power. Also, episodes were limited to around 15,000, which is especially unideal for measuring specific metrics for PPO in dynamic environments. PPO and DQN would benefit from extended training and larger batch sizes. However, significantly more episodes will be tested for future research, as some research shows training has ran for 200,000 plus episodes. Additional algorithms and episodes would be an excellent experimental choice, as each gives a richer comparison and could lead to interesting and more precise findings, especially a hybrid Q-learning algorithm. By incorporating a hybrid, Q-learning can increase its learning capabilities and make better comparisons against algorithms with networks. Further improvements using other types of metrics should also be considered. Could the entropy for PPO and DQN be visualised across time to explore conversion trends? And could the agent's success rate be compared with known benchmarks for real guide dogs? These questions open many opportunities for further study and could better correlate this paper to real-life findings. The next steps would include future research, in which a minimum of 50,000 episodes should be simulated, as preliminary results suggest this for optimal performance. Access to GPU clusters or distributed training frameworks would allow scaling to more extensive experiments in future work. This would hopefully lead to better experimentation and results. Finally, the ultimate question should be answered: Evaluating reinforcement learning algorithms for navigation in simulated robotic quadrupeds: A comparative study inspired by guide dog behaviour.

Although experimentation was limited, I believe quadrupedal robots can perform equally as well as thoroughly trained guide dogs. With more time and greater training, I do believe that statistically, robots will be able to perform equally as efficiently as a trained guide dog.

# Appendices

**Appendix 1.**

Fig. 1. A simple environment built in Webots (left), for the quadruped to explore, with LiDAR activated. A Dynamic environment (right) for the quadruped to explore

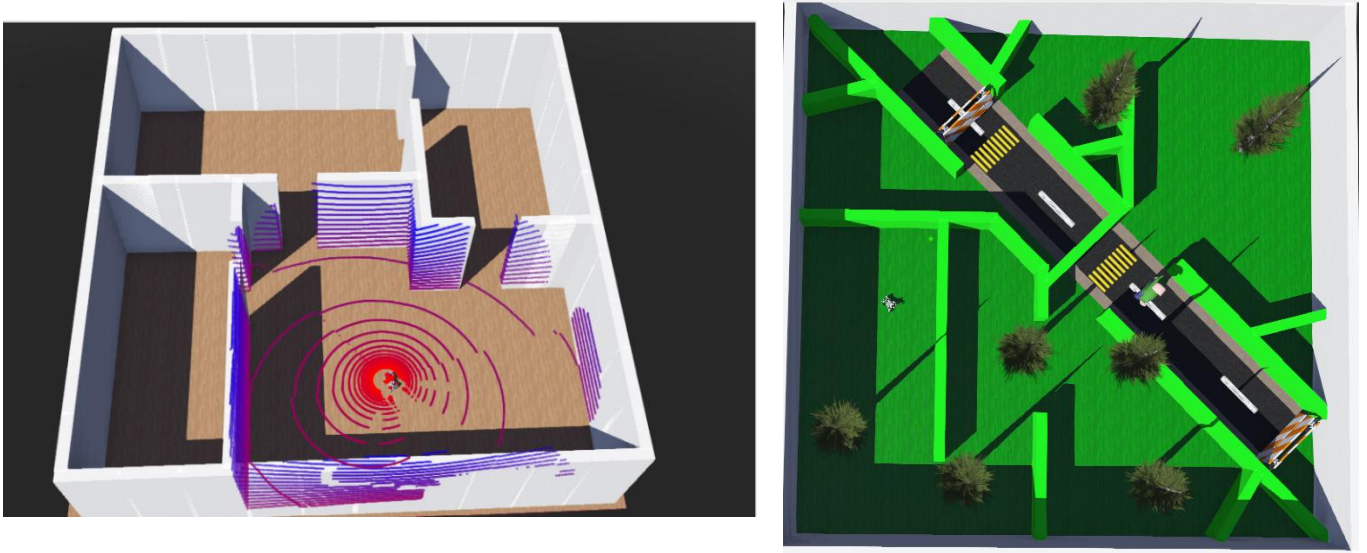

**Appendix 2.**

Fig. 2. Shows a flowchart of the experimental procedure for visual ease.

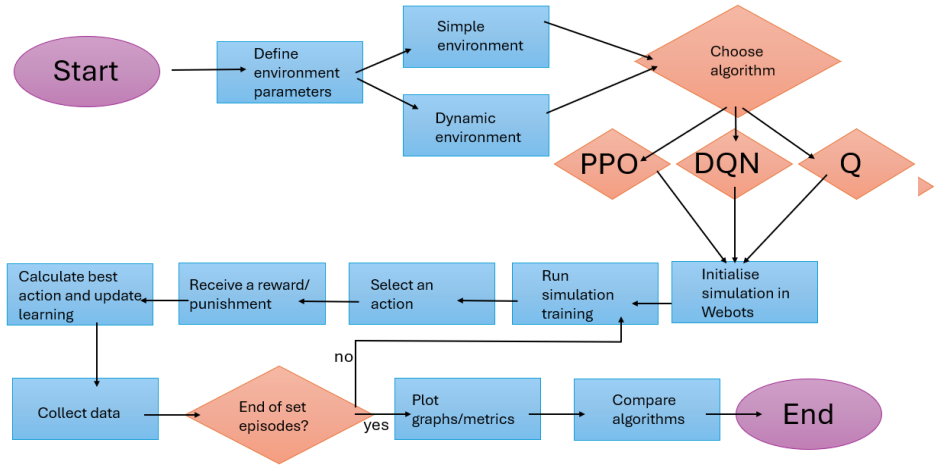

**Appendix 3.**

Fig. 3. Shows a snippet of code in which the main controller talks to the supervisor controller asking to reset the quadruped after a fall.

```
def sendToSupervisor(condition: str, value: bool): #fallen, goalReached, timeout):
    #Send JSON data to supervisor for failure / reset conditions
    #TODO: Input datatypes?
    jsonData = {
                "fall": (condition == "fallen" and value == True),
                "goal": (condition == "goal" and value == True),
                "time": (condition == "timeout" and value == True)
            }
    socketOUT.send_string(json.dumps(jsonData))
```

**Appendix 4.**

Fig. 4. Graphs to show reward per episode with smoothing for PPO, DQN, and Q. Results for the simple environment on the left and the dynamic environment on the right.

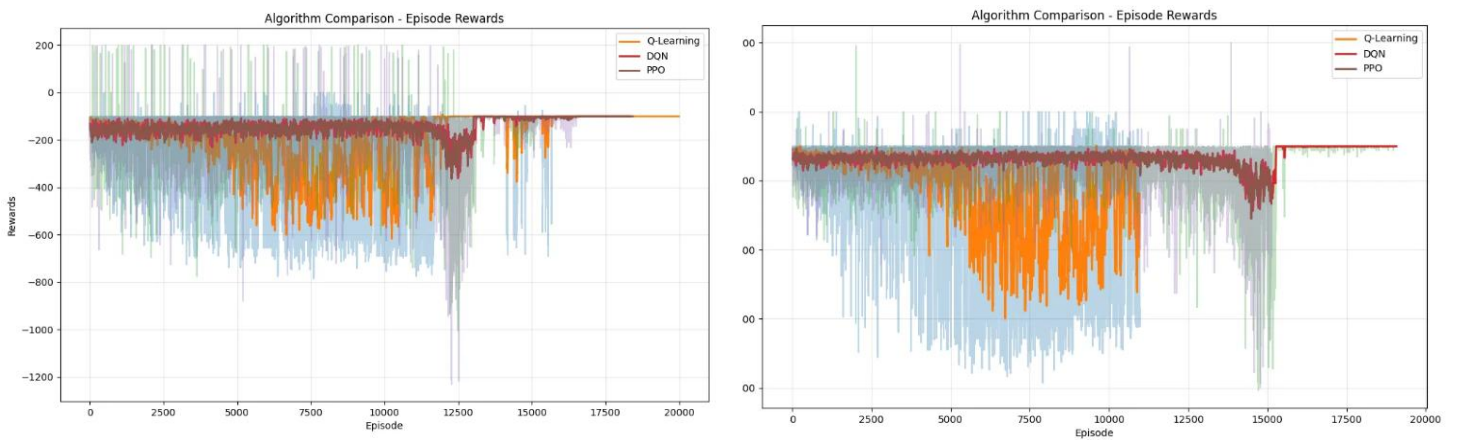

**Appendix 5.**

Fig. 5. A graph to show the learning rate between PPO, DQN, and Q.

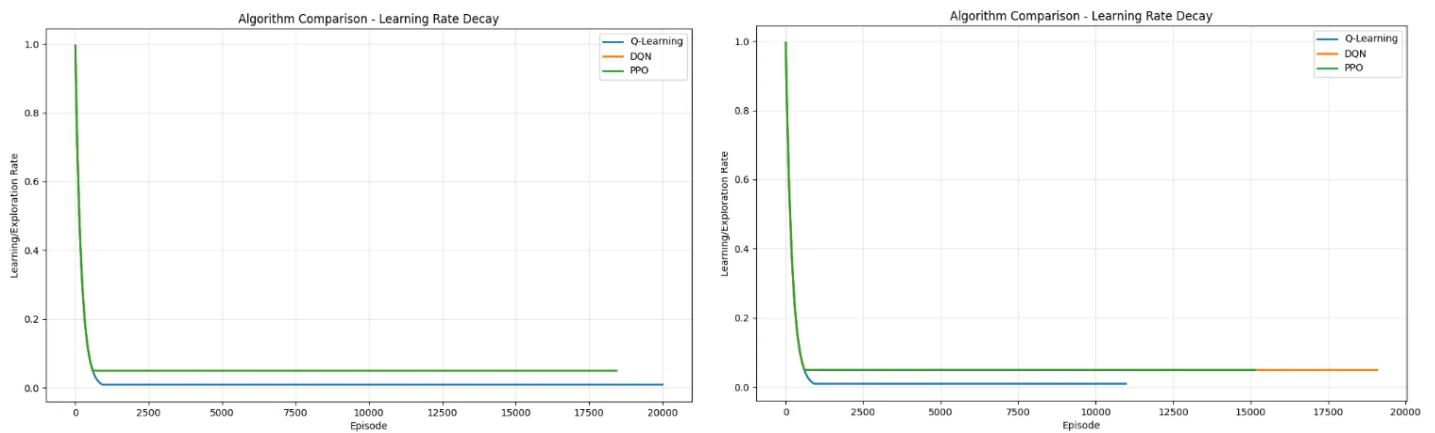

**Appendix 6.**

Fig. 6. Graph showing the number of steps per episode between PPO, DQN, and Q. The Goal reached is represented by stars. The performance results from a simple environment are on the left, and the performance results from a dynamic environment are on the right.

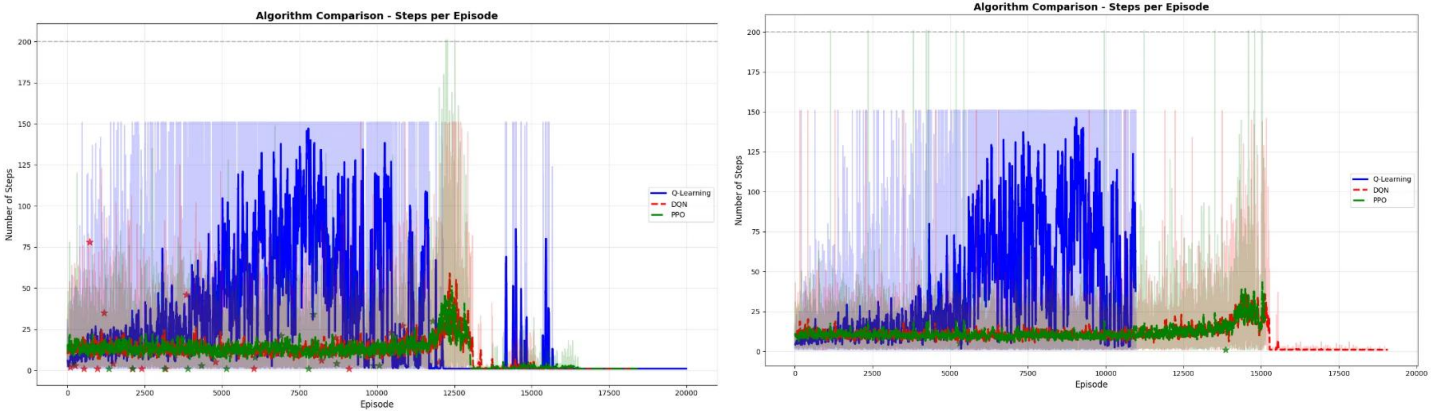

**Appendix 7.**

Fig. 7. A graph to show the average and median steps taken to the goal with success rate per episode for PPO, DQN, and Q. Left shows metrics for the simple environment, right for the dynamic environment.

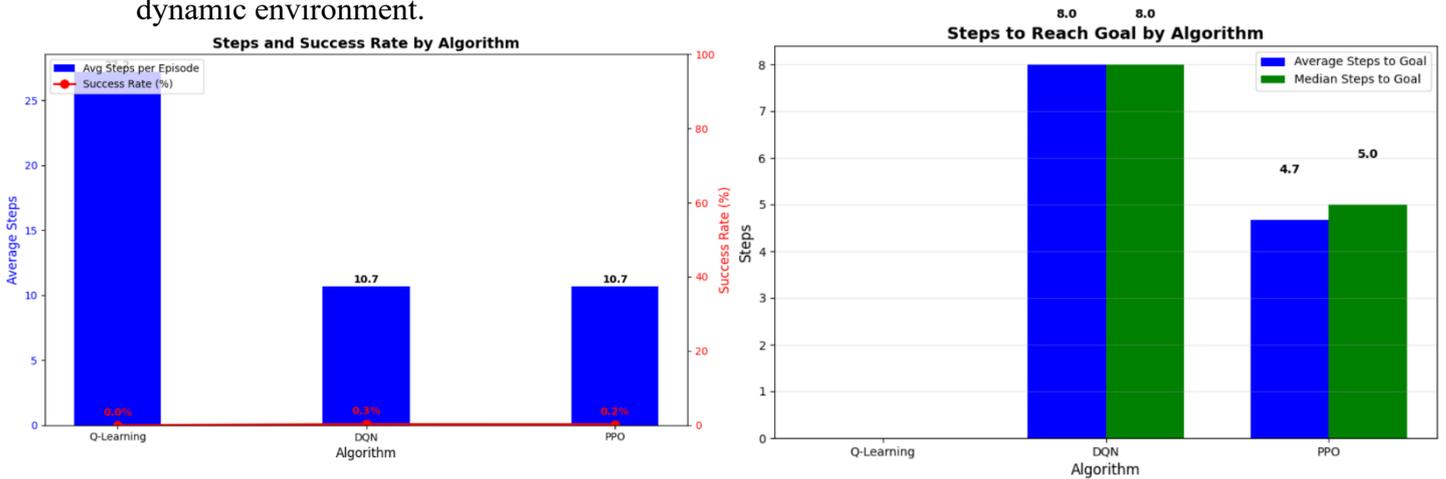

**Appendix 8.**

Fig. 8. Shows the number of collisions between PPO, DQN and Q with smoothing. A simple environment results in the left, and dynamic environment results are displayed on the right.

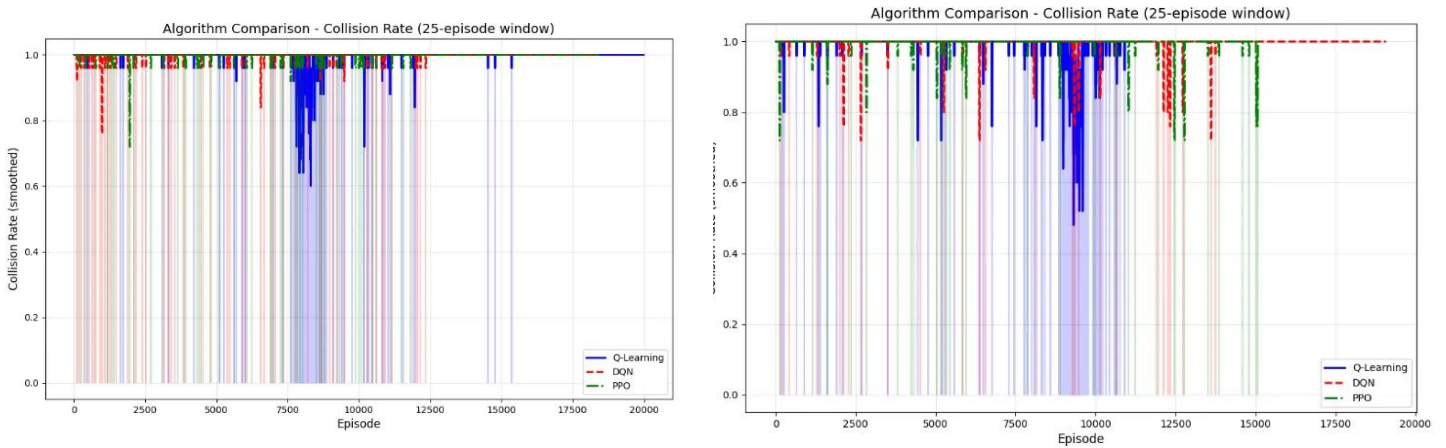

Fig. 9. Shows the statistical analysis of the difference in means of average reward per episode by each algorithm and the difference in means of average steps per episode by each algorithm in a simple environment.

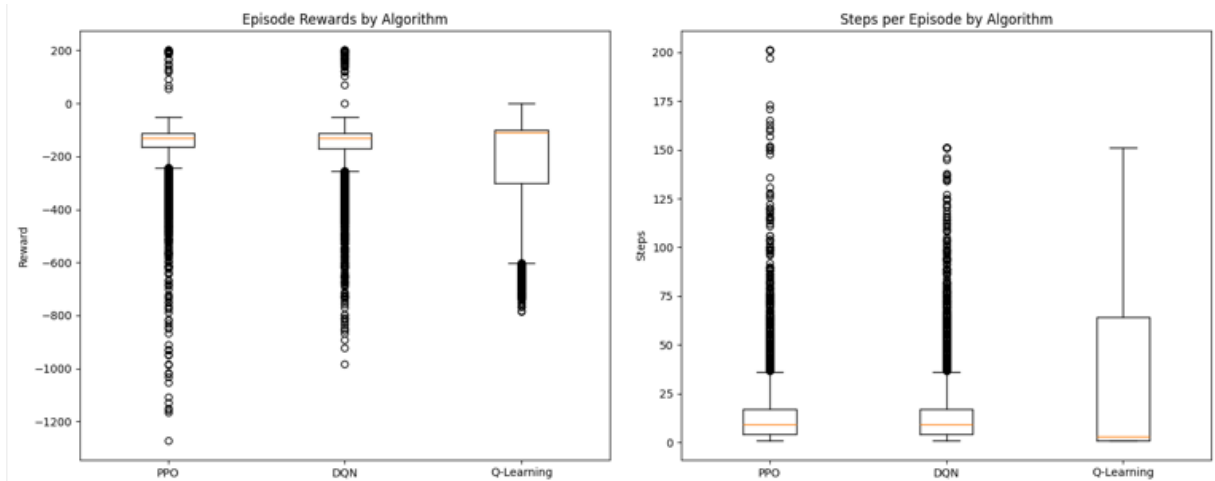

Fig. 10. Shows the statistical analysis of the difference in means of average reward per episode by each algorithm and the difference in means of average steps per episode by each algorithm in a dynamic environment.

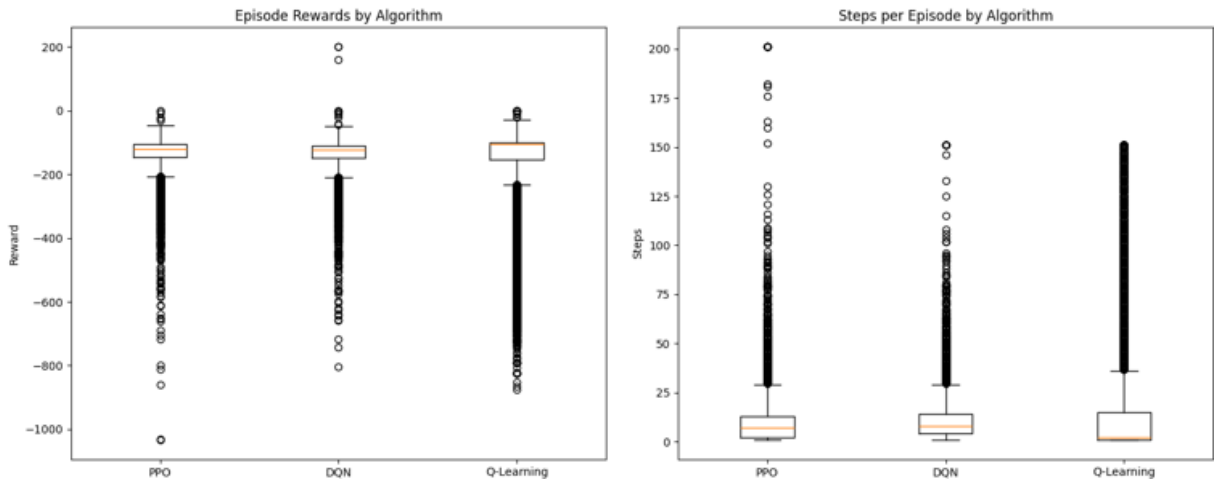